\documentclass{article}

\PassOptionsToPackage{numbers,sort&compress}{natbib}
\usepackage[final,dblblindworkshop]{neurips_2026}
\ifdefined\workshoptitle
  \workshoptitle{Real-Time Conversational Agents: Toward Natural Multimodal Interaction}
\fi

\usepackage[utf8]{inputenc}
\usepackage[T1]{fontenc}
\usepackage[hidelinks]{hyperref}
\usepackage{url}
\usepackage{booktabs}
\usepackage{amsfonts}
\usepackage{amsmath}
\usepackage{amssymb}
\usepackage{microtype}
\usepackage{graphicx}
\usepackage{xcolor}
\usepackage{wrapfig}
\usepackage{placeins}
\usepackage{enumitem}
\usepackage{array}
\usepackage{float}

\newcommand{\bench}{\textsc{PeopleMakingConversation}}
\newcommand{\tb}{\textsc{TurnBench}}
\graphicspath{{figures/}}

\title{People Making Conversation:\\Evaluating How Naturally Speech Models Converse}

\author{%
  Arya Bastani\thanks{These authors contributed equally.} \\
  People Make Things \\
  \texttt{arya@peoplemakethings.com} \\
  \And
  Iman Modarressi\footnotemark[1] \\
  People Make Things \\
  \texttt{i@peoplemakethings.com} \\
}

\begin{document}

\maketitle
\begin{abstract}
A conversational speech model has to decide, at every pause, whether to speak. Existing evaluations test boundary detection, policy adherence, or aggregate statistics, not this decision. We place real-time speech systems at 427 pauses and 482 turn ends from 176 unscripted conversations in Seamless Interaction, give each the conversation so far, and score what it generates against what the person did: whether and how soon it speaks at a pause; its latency, length, and wording at a turn end; and whether it can answer a question about the preceding 30 to 120 seconds. Across six systems, the hosted APIs (GPT Realtime, Gemini 3.1 Flash Live, Grok Voice) respond at almost every pause whichever endpointer they use; only their latency, about 1 to 4\,s with silence-based endpointers and about 8\,s with semantic turn detection, decides whether the words land inside the pause. The full-duplex models (PersonaPlex, Moshi) are the only systems that regularly stay silent and the only ones that interrupt within a human reaction time. Hosted systems reply longer than the person. Given the history as text, hosted systems recall 41 to 60\% of conversation facts under a Claude Opus 5.5 judge and 19 to 26\% under a strict word matcher; hearing it as audio, full-duplex models recall 17 to 22\% and 12 to 15\%. We release the labels, points, harness, and every output.\footnote{\url{https://github.com/People-Make-Things/people-conversation-bench}}
\end{abstract}

\section{Introduction}

A conversational model must know when to say nothing. A listener can take
the floor, acknowledge without taking it, or keep listening. The choice
depends on the conversation so far, not only on whether the other person
has stopped speaking. A short response latency does not establish whether the system should have spoken. Natural here means doing what the person did at that moment: silent where they were silent, replying where they replied, with similar timing and length; no listener rates the outputs.

\begin{figure}[!t]
\centering
\includegraphics[width=\linewidth]{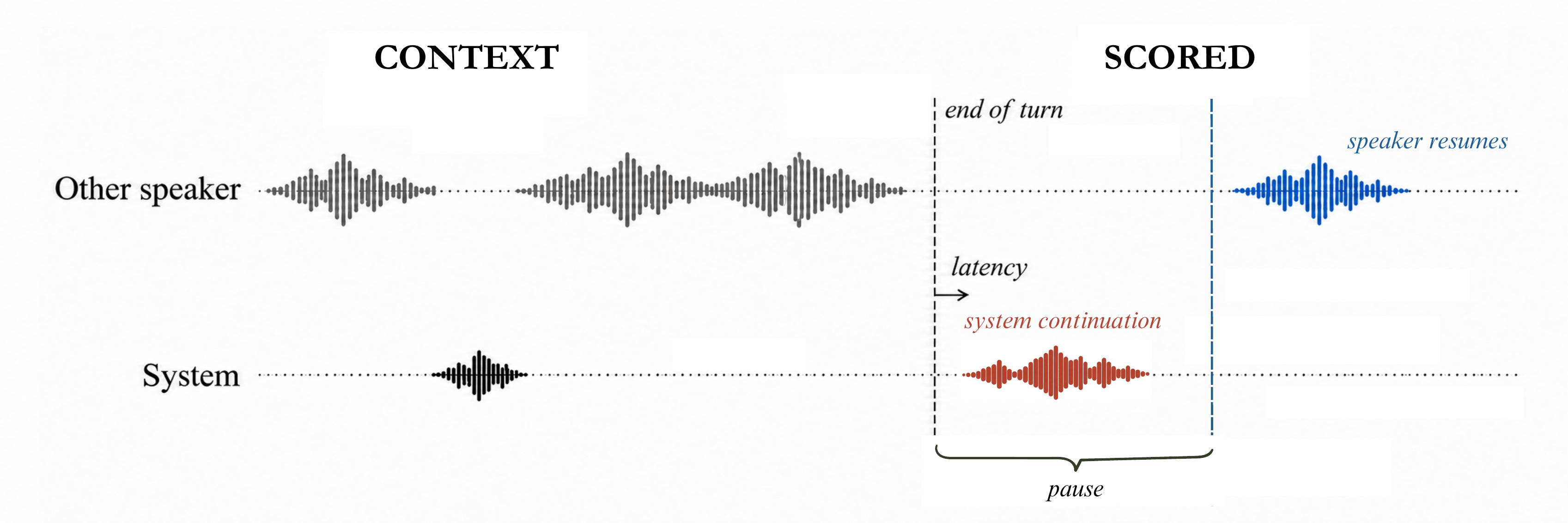}
\caption{A pause point. The system is preloaded with the conversation up to
the active turn (context), then hears the active turn live. The dashed line
is where the speaker stops, which a system may take for the end of the
turn; the pause runs until the speaker resumes (blue). A system that
produces words inside the pause has taken the floor (red); latency is
measured from the pause start to its first words. Only the scored region
counts. Turn-end points end the live stream at the turn end and score the
response instead.}
\label{fig:protocol}
\end{figure}

\bench\ does that. From 176 casual conversations we select two kinds of
moment. A \emph{pause point} is a silence inside a speaker's turn that a
system might mistake for its cue; the person listening did not speak, and
the speaker went on. A \emph{turn-end point} is a boundary where the floor
passed and the person replied. The system receives the conversation up to
the moment, then hears the other speaker continue in real time
(Figure~\ref{fig:protocol}). At a pause we score whether it stays silent or
takes the floor, and how quickly. At a turn end we score how soon it begins,
how long it speaks relative to the person, and how close its words are to
theirs. After 30 to 120 seconds of history we also ask a spoken question about something said earlier, and a judge scores the answer.

We evaluate six real-time systems, two self-hosted full-duplex models and
four hosted API configurations. Two things follow from the results. With the silence-based endpointers we tested, hosted systems almost always responded; their latency, 1 to 4\,s, rather than a contextual turn-taking decision, determined whether speech began inside the pause. The full-duplex models are the only
systems that regularly stay silent, and the only ones that interrupt within a
human reaction time; at turn ends they do not stop, and no system says as
little as the person did. We release the corpus selection, labels,
evaluation points, the harness, and every system output at \url{https://github.com/People-Make-Things/people-conversation-bench}.

\section{Related work}
\label{sec:related}

Moshi reports aggregate timing statistics \citep{moshi}. The Gemini 3.1 Flash Live and GPT-Live releases report audio reasoning, function calling, and human preference, but no boundary-level account of which decisions the model gets right \citep{geminilive,gptlive}. Model papers such as BayLing-Duplex and SALMONN-omni score scenario success \citep{bayling,salmonn}.

Table~\ref{tab:corpora} lists the corpora of two-party conversation and
Table~\ref{tab:benchmarks} the benchmarks. Turn-taking is measured in three
ways.
\emph{Detection} benchmarks ask whether a system marks the events a human
annotator marked: end of turn, pause, backchannel, interruption. \tb,
Talking Turns, and Full-Duplex-Bench v1 do this, and each penalises firing
at a pause within a turn \citep{turnbench,talkingturns,fdb1}. \emph{Policy}
benchmarks ask whether a system speaks when a rule says it should, or
completes a task while conversing: Speak or Stay Silent, CoDeTT,
Instruct-FD, Full-Duplex-Bench v2/v3, and MTR-DuplexBench
\citep{speakorstay,codett,instructfd,fdb2,fdb3,mtrduplex}. \emph{Aggregate}
evaluations ask whether generated dialogue resembles human dialogue in its
statistics, or whether people prefer it \citep{syncllm,gptlive}. Each
answers its own question; none asks ours.

\begin{table}[H]
\caption{Corpora of two-party conversation. Hours are approximate.}
\label{tab:corpora}
\centering
\footnotesize
\setlength{\tabcolsep}{5pt}
\begin{tabular}{@{}lrllc@{}}
\toprule
Corpus & Hours & Setting & Spontaneous & Turn-taking labels \\
\midrule
Switchboard \citep{switchboard} & 260 & telephone & yes & turns only \\
Fisher \citep{fisher} & 1{,}960 & telephone & yes & -- \\
CANDOR \citep{candor} & 850 & video call & yes & -- \\
SSSD \citep{sssd} & 727 & crowdsourced & yes & -- \\
CASPER \citep{casper} & 200 & crowdsourced & yes & -- \\
Seamless Interaction \citep{seamless} & $>$4{,}000 & in person, one channel per speaker & naturalistic subset & ours (\S\ref{sec:labels}) \\
\tb\ corpus \citep{turnbench} & 30 & studio, voice actors, assigned type & no & triple-annotated \\
\bottomrule
\end{tabular}
\end{table}

\begin{table}[H]
\caption{Benchmarks for turn-taking in speech systems. Continuation: the
generated continuation is scored against what the human did at the same
boundary.}
\label{tab:benchmarks}
\centering
\footnotesize
\setlength{\tabcolsep}{3.5pt}
\begin{tabular}{@{}>{\raggedright\arraybackslash}p{2.75cm}l>{\raggedright\arraybackslash}p{2.55cm}>{\raggedright\arraybackslash}p{4.0cm}c@{}}
\toprule
Benchmark & Kind & Data & What is scored & Contin. \\
\midrule
Full-Duplex-Bench v1 \citep{fdb1} & detection & synthetic probes, CANDOR, ICC & pause takeover, backchannel rate and timing, latency, interruption & -- \\
Talking Turns \citep{talkingturns} & detection & Switchboard subset & backchannel, end of turn, turn change, by judge model & -- \\
\tb\ \citep{turnbench} & detection & own actor corpus & end of turn and interruption: recall, FPR, latency & -- \\
\addlinespace[2pt]
Speak or Stay Silent \citep{speakorstay} & policy & multi-party transcripts & speak vs.\ silent at each pause, from text & -- \\
CoDeTT \citep{codett} & policy & LLM-scripted, TTS-rendered scenarios & context-conditioned decision categories & -- \\
Instruct-FD \citep{instructfd} & policy & synthetic instruction-conditioned dialogues & adherence to a stated turn policy, by LLM judge & -- \\
Full-Duplex-Bench v2/v3 \citep{fdb2,fdb3} & task & LLM examiner; disfluent audio & turn fluency, task success, tool calls & -- \\
MTR-DuplexBench \citep{mtrduplex} & task & multi-round dialogue & fluency, instruction following, safety & -- \\
\addlinespace[2pt]
SyncLLM eval \citep{syncllm} & aggregate & human continuations & corpus-level correlation of pauses, backchannels, turn length & aggregate \\
VoiceBench \citep{voicebench} & content & spoken QA & answer correctness & -- \\
\midrule
\textbf{Ours} & continuation & 176 naturalistic conversations & hold vs.\ takeover at pauses; latency, length, content at turn ends; recall vs.\ history & \checkmark \\
\bottomrule
\end{tabular}
\end{table}

The detection family is the nearest to ours. What it scores is whether a boundary event was emitted at the
right time. What we score is what a generative system does at a boundary a
person actually faced, in a conversation nobody staged, and what it retains
from the minutes before. Their stimuli are elicited or synthetic by design,
which buys controlled conditions; ours are casual by design, which buys
the decisions we want to measure. The two are complementary, and a system
should be run on both. Our hosted pause results differ from benchmarks where hosted systems appear to hold at pauses. Those evaluations let the provider's turn detector decide when a turn ended and score the detection; we requested a response once a silence-based endpointer fired, so what we measure is that configuration's behaviour and latency, not whether the model could hold under a different turn detector (\S\ref{sec:limitations}).

\section{The benchmark}
\label{sec:bench}

\subsection{Corpus}
\label{sec:corpus}

The corpus is the naturalistic subset of Seamless Interaction
\citep{seamless}: 1{,}395 prompted, casual sessions of at least seven
minutes, in which two participants were given a topic card and talked. We
select by liveliness alone. Six traits are computed from the released
transcripts and voice activity (speech-time balance, speaker alternations
per minute, short-utterance rate, question rate, one minus the longest
single-speaker share, overlaps per minute); a conversation's score is the
mean of its six percentile ranks among the 1{,}395. The 176 at or above 0.75 are the corpus
(35.9 hours, mean 12.2 minutes, median 8.7; Appendix~\ref{app:selection}).
No model selects conversations; the LLM judge of \S\ref{sec:points} selects
points within them. The score favours lively conversation, so rates
reported here are not population estimates.

\subsection{Labels}
\label{sec:labels}

Each speaker's channel is labelled with speech onsets and offsets, turns,
pauses, backchannels, and overlaps. Voice activity and transcripts are the
corpus's own; voiced stretches of at least 350\,ms with no transcript are
transcribed and aligned by Qwen3-ASR and Qwen3-ForcedAligner. A
deterministic rule engine places every label and records the rule and
confidence behind it. Gaps of at most 250\,ms inside one speaker's activity
are closed. A listener segment of at most 1\,s and two words during the
other speaker's turn is a backchannel. At every offset the engine looks
ahead 5\,s: if the same speaker resumes it is a pause; if the other speaker
starts substantively it is an end of turn; if the other speaker starts and
yields it is a pause and a failed take; if nobody speaks it is an end of
turn. A voice activity projection model \citep{vap} and three heuristics
flag labels for review: a turn ended while the projection keeps the speaker
on the floor ($P \geq 0.65$), a backchannel where it hands the floor over
($P \geq 0.60$), a turn with half its words from gap-fill, and a lexical
turn under 400\,ms; 19{,}933 turns produce 5{,}447 flags (27.3\%,
Appendix~\ref{app:mapping}). The points in this paper are built on the
engine's labels. Every pause and turn-end window in the run was played with
its context to two annotators and accepted, an accept-or-reject check with
no per-window log.

\subsection{Evaluation points}
\label{sec:points}

The 176 conversations contain about 46{,}000 candidate boundaries; the point set is curated from them in two stages. Hard gates reject events that
cannot be scored cleanly: pauses shorter than 600\,ms or overlapped by the
listener, active turns shorter than 1\,s or three words, turn ends whose
reply starts more than 3\,s later or has fewer than five words, and events
whose audio fails the scorer's speech gate. Survivors are ranked by a
heuristic over pause length, turn context, and the last words before the
boundary, keeping at most three per type per conversation. A GPT-4o judge
then reads a transcript window around each survivor. It rejects incoherent
windows, pauses that hand the turn to the listener, interrupted fragments,
and unresponsive replies, and ranks the rest by how tempting the pause is.
An event the judge scores zero is never selected. This gives 427 pause
points and 482 turn-end points across 157 and 171 of the 176 conversations;
the run reported here draws a seeded random 300 of each. At a selected
pause the engine found no listener words and at a selected turn end the
person replied substantively, so the human reference is hold or take,
never backchannel. Appendix~\ref{app:selection} gives the funnel.

Recall points test memory. For each prefix of 30, 60, 90, and 120 seconds
ending at a completed turn, a question is spoken to the system in the
user's voice, cloned by Qwen3-TTS from 20 to 40\,s of their speech.
\emph{Conversation recall} asks, in the user's words and addressed to
the system, about a specific fact either speaker stated in the first 30
seconds. The system plays speaker 1; the question is spoken in speaker 2's cloned voice, with ``I'' and ``me'' referring to speaker 2 and ``you'' to speaker 1. A correct answer states the fact from speaker 1's position, whichever speaker originally said it (example in Table~\ref{tab:exrecall}). GPT-4o extracts it, a judge checks it for specificity, and it is
dropped if its spoken form leaks a speaker label. Answers are scored by Claude Opus 5.5 (temperature 0); GPT-4o scores of the same transcripts are kept for comparison (Table~\ref{tab:judge}). The same question is asked
at every history length. \emph{Fact recall} is the control. It asks
one general-knowledge question from a fixed list of twelve, assigned per
conversation, whose answer does not depend on the history; a system that
fails it did not answer the question it was asked. A judge-approved
conversation fact exists for 46 conversations, giving 184 conversation-recall
points; fact recall has 669, a prefix being skipped when no turn completes
before its target. The run uses all 184 and a seeded random 300.

\subsection{Protocol}
\label{sec:protocol}

A rollout has two phases: history preload, then a live stream that is
scored. History is capped at 164\,s before the active turn, PersonaPlex's
context window. Full-duplex models receive it as two audio channels, the
other speaker's and their own, teacher-forced with word timestamps. Hosted
models receive completed turns as text and the trailing user audio; their
APIs accept nothing else. Every system also receives a short identity
prompt, written once per speaker by GPT-4o from the transcript: who it is,
how it knows the other person, and that the assistant history is its own
prior speech.

At a pause point every system is streamed the active turn, the recorded
pause, and the speaker resuming, as real-time frames, so it hears the pause
as real silence; a full-duplex model steps through it frame by frame, and a
hosted model's server-side endpointer fires on it. The endpointer is the
system's turn-taking policy. At a turn-end point the trigger differs by family: a hosted model receives an explicit end-of-turn signal and a requested response, with its endpointer off so a mid-turn pause cannot interrupt it; a full-duplex model must detect the end from the audio itself and then receives silence frames until it stops or reaches the cap. The person's reply is not played back. Drain lengths, caps, and audio format are in
Appendix~\ref{app:cards}. We score the first rollout of each point; a second rollout from the same run is released but not reported. Every output is
recorded with its input timing, so every score can be recomputed from the
rollout.

\subsection{Metrics}
\label{sec:metrics}

Speech intervals are found by an energy gate calibrated against Whisper
word timestamps on the systems' own output (Appendix~\ref{app:sens}). The
gate alone is a weak classifier of speech, so a \emph{response} requires
transcribed words; above-gate audio with an empty transcript is
\emph{wordless phonation}, reported as its own category.

\begin{enumerate}[leftmargin=1.4em,itemsep=1pt,topsep=2pt,parsep=0pt]
\item \textbf{Pause recognition.} On the stream of \S\ref{sec:protocol}, a
\emph{takeover} is a word-bearing interval overlapping the labelled pause;
otherwise the system \emph{held}, either silently or by speaking after the
person resumed. We report the shares silent, takeover, and after, the
wordless rate, and \textbf{takeover latency} from pause start to the first
words, zero if the system was already speaking.
\item \textbf{Turn-end latency.} Time from the turn end to the onset of the
system's first transcribed word, on the playback clock; a word that began
before the turn end gives zero or less.
\item \textbf{Response length.} Duration from first to last speech interval
and word count, each as a ratio to the human reply.
\item \textbf{Response similarity.} Token F1 and cosine similarity of
\texttt{all-MiniLM-L6-v2} embeddings between the system's transcript and the
human reply.
\item \textbf{Recall.} Whether the spoken answer is correct, judged by
Claude Opus 5.5 against the gold answer or the conversation, reported by history
length.
\end{enumerate}

\section{Experiments}
\label{sec:experiments}

\subsection{Systems}

We evaluate six real-time speech systems in two families. Self-hosted
full-duplex models listen and speak at once and run on Modal: PersonaPlex-7B
\citep{personaplex} and Moshi \citep{moshi}. Hosted real-time
APIs take audio in and return audio out, with the provider's server-side
voice activity detection deciding when the user's turn has ended: GPT
Realtime (\texttt{gpt-realtime-2025-08-28}) \citep{gptrealtime}, Gemini 3.1
Flash Live (preview) \citep{geminilive}, and Grok Voice
(\texttt{grok-voice-latest}) \citep{grokvoice}. GPT Realtime is
run at two silence settings, 300\,ms and 500\,ms, which isolates the
endpointer's contribution within one model. The 500\,ms setting is OpenAI's documented server-VAD default; 300\,ms is the faster variant \citep{vad_docs}. For pause points only, GPT Realtime is also run with OpenAI's semantic turn detection at eagerness \texttt{low}, the setting most willing to wait; turn-end and recall evaluations turn detection off and commit the turn explicitly, so this configuration has no results there. GPT Realtime and Grok use a
detection threshold of 0.5, Gemini's automatic detection closes on 500\,ms
of silence, and every hosted endpointer uses 300\,ms of prefix padding.
Versions,
evaluation dates, and settings for every system are in
Appendix~\ref{app:cards}.

\subsection{What is held fixed}

Every system sees the same points, the same history cap, the same identity
prompt, the same live audio frames, and the same speech gate, and is scored
by the same code. One thing differs and is not hidden: full-duplex models hear the history and hosted models read it, so cross-family recall comparisons carry that caveat (\S\ref{sec:limitations}).

\subsection{Reference points}

Two policies bound the pause and turn-end metrics without running
anything: never speak holds every pause and takes no turn; always speak
takes every pause at zero latency. The human, by construction of the point
set, held every pause and took every turn end, with a length ratio of 1 and
a similarity of 1. Similarity has a measured floor, the human reply from
a different point in the same conversation (median token F1 0.08, cosine
0.12 over 240 pairs), and a measured ceiling, the human reply against its
own label transcript (0.88, 0.89), the transcription noise every system is
scored through. Recall has
a ceiling too: a text model reading the transcript answers 97\% of the
conversation-recall questions under the GPT-4o judge (Appendix~\ref{app:extended}), so a score is
a fact about memory, not about the questions.

\subsection{Reporting}

Every metric is reported per system and per point type, never pooled across types. Recall is pooled over history lengths only for the scorer comparison (Table~\ref{tab:judge} and the abstract). The semantic configuration ran on 475 curated pauses, two rollouts each; comparisons with the 300-point set use the 173 shared points. A
rollout that returns no audio or times out is an outcome reported beside
the scores, not an exclusion. Rates are over rollouts; latencies, ratios,
and similarities are medians with interquartile ranges over rollouts with
words; confidence intervals (Appendix~\ref{app:extended}) come from a
bootstrap over conversations, which share a speaker, a topic, and a
recording. We report the metric vector and do not combine it.

\section{Results}
\label{sec:results}

\begin{figure}[t]
\centering
\includegraphics[width=\linewidth]{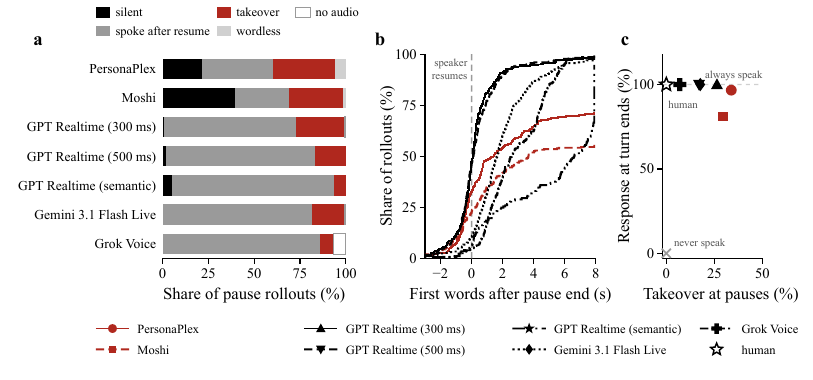}
\caption{Pause points. (a) What each system did at the person's pause;
the person stayed silent, so more black is better. (b) When each system's
first words landed relative to the pause end (dashed line); the hosted
curves all reach the top because every hosted rollout ends with the system
speaking; red is full-duplex, black is hosted. (c) Each system as one point: takeover rate at pauses against
response rate at turn ends; symbols as in the index below. The person is the open star at the top left; the
hosted systems sit on the always-speak line. GPT Realtime (semantic) is the low-eagerness semantic turn-detection run on the 173 shared pause points; it has no turn-end result, so it is absent from (c).}
\label{fig:pause}
\end{figure}

\textbf{Pause points} (Figure~\ref{fig:pause}, Table~\ref{tab:main}). The hosted systems almost never stayed silent (0 to 2\% of points): their silence-based endpointers request a response once speech ends. Silent shares and no-audio outcomes are reported separately (Table~\ref{tab:agreement}). Hosted speech began a median 1.0 to 1.1\,s after the pause began for GPT Realtime,
2.7\,s for Gemini, and 3.9\,s for Grok over all rollouts (Table~\ref{tab:main}
gives the median over takeovers only), against a median labelled pause of
0.97\,s. Whether the words land inside the pause, a takeover on 7 to 26\%
of points, or after the person has resumed, on 72 to 86\%, is set by the
latency, not by a decision. Grok's low takeover rate is its slowness, not
restraint: its words arrive after the person has resumed on 86\% of points.
The full-duplex models are the only systems that stay silent, on 22\%
(PersonaPlex) and 39\% (Moshi) of points, because nothing forces a full-duplex model to respond. They are also the only
ones that take the floor within a human reaction time, a median 216 and
373\,ms in; three in ten of PersonaPlex's takeovers and a quarter of
Moshi's were speech already under way at the pause. Raising GPT Realtime's endpointer from 300 to 500\,ms cuts its
takeover rate from 26 to 17\% and delays its speech by 80\,ms: the
endpointer is the policy. Semantic turn detection at low eagerness changes the timing more than the outcome: on the 173 pauses shared with the main set it produced no words on 5\% (9 points, several with wordless audio), took the floor inside the pause on 6\% (500\,ms: 17\%; 300\,ms: 26\%), and spoke after the person resumed on 88\%, its first words a median 6\,s after the pause ended (Table~\ref{tab:main}, Figure~\ref{fig:pause}). Wordless phonation is a full-duplex phenomenon,
6\% of PersonaPlex's points and 1\% of Moshi's.

\begin{table}[b]
\caption{Main results, one rollout per point. The \textit{better} row gives
the good direction and the \textit{person / ceiling} row the target: the
person stayed silent at every pause and replied at every turn end, and the
ceilings are what a perfect answer scores through transcription and judging
noise. Bold is the best system per column. Pause: share of points silent,
taking the floor inside the pause, or speaking after the person resumed, and
median takeover latency. Turn end: latency to first word, share still
speaking at the 30\,s cap (cens.), duration and word ratios to the human
reply, token F1 and cosine. Recall: accuracy at 120\,s, Claude Opus 5.5 judge; the 0.97 ceiling is GPT-4o-judged. Moshi's latency is
speech already under way; Grok's includes a 0.5\,s protocol wait. Intervals
in Table~\ref{tab:ci}. $^{\dagger}$Semantic turn detection ran on pause points only; its row uses the 173 pause points shared with the main set.}
\label{tab:main}
\centering
\footnotesize
\setlength{\tabcolsep}{3pt}
\resizebox{\linewidth}{!}{\begin{tabular}{@{}l r r r r r r r r r r r r@{}}
\toprule
 & \multicolumn{4}{c}{Pause points} & \multicolumn{6}{c}{Turn-end points} & \multicolumn{2}{c}{Recall@120\,s} \\
\cmidrule(lr){2-5}\cmidrule(lr){6-11}\cmidrule(lr){12-13}
System & silent & takeover & after & lat.\ (ms) & lat.\ (ms) & cens. & dur.\ ratio & word ratio & token F1 & cosine & conv. & fact \\
\textit{better} & $\uparrow$ & $\downarrow$ & $\downarrow$ & -- & $\downarrow$ & $\downarrow$ & $\approx 1$ & $\approx 1$ & $\uparrow$ & $\uparrow$ & $\uparrow$ & $\uparrow$ \\
\midrule
\textit{person / ceiling} & 1.00 & 0.00 & 0.00 & -- & -- & 0.00 & 1.00 & 1.00 & 0.88 & 0.89 & 0.97 & 1.00 \\
\midrule
PersonaPlex & 0.22 & 0.34 & 0.38 & 216 & 1120 & 0.90 & 7.81 & 3.87 & 0.10 & 0.16 & 0.13 & 0.23 \\
Moshi & \textbf{0.39} & 0.29 & \textbf{0.30} & 373 & \textbf{86} & 0.92 & 3.38 & 0.77 & 0.00 & 0.12 & 0.17 & 0.25 \\
\addlinespace[2pt]
GPT Realtime (300 ms) & 0.01 & 0.26 & 0.72 & 983 & 514 & 0.03 & 3.63 & 3.00 & \textbf{0.12} & 0.19 & 0.47 & \textbf{0.97} \\
GPT Realtime (500 ms, default) & 0.02 & 0.17 & 0.81 & 1065 & 479 & 0.02 & 3.54 & 3.00 & \textbf{0.12} & \textbf{0.21} & 0.46 & \textbf{0.97} \\
GPT Realtime (semantic, low)$^{\dagger}$ & 0.05 & 0.06 & 0.88 & 1339 & -- & -- & -- & -- & -- & -- & -- & -- \\
Gemini 3.1 Flash Live & 0.00 & 0.18 & 0.82 & 2355 & 1672 & 0.24 & 3.28 & 2.61 & \textbf{0.12} & 0.18 & 0.38 & 0.62 \\
Grok Voice & 0.00 & \textbf{0.07} & 0.86 & 3045 & 922 & \textbf{0.01} & \textbf{1.39} & \textbf{1.14} & \textbf{0.12} & 0.18 & \textbf{0.61} & 0.87 \\
\bottomrule
\end{tabular}
}
\end{table}

\begin{figure}[t]
\centering
\includegraphics[width=\linewidth]{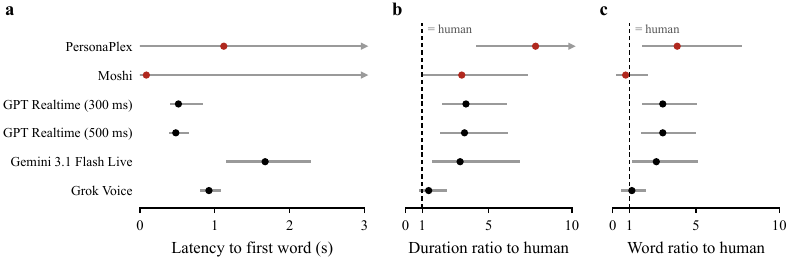}
\caption{Turn-end points. (a) Latency to first word. (b) Response duration
and (c) word count as ratios to the human reply; the dashed line is the
human. Median and interquartile range over rollouts with words, linear
axes; an arrowhead marks an interquartile range that runs past the axis,
and a latency at zero is a word that began before the turn end.}
\label{fig:eot}
\end{figure}

\textbf{Turn-end points} (Figure~\ref{fig:eot}). GPT Realtime answers a
median 0.5\,s after the turn ends, Grok 0.9\,s (of which 0.5\,s is a
protocol wait), PersonaPlex 1.1\,s, Gemini 1.7\,s. Moshi's median is
86\,ms because it was often already talking through the turn; on 19\% of
points it produced no words at all. Every system talks longer than the
person, in seconds if not always in words. Among replies with words, the
median duration ratio is 1.4 for Grok, 3.3 to 3.6 for Gemini, Moshi, and
GPT Realtime, and 7.8 for PersonaPlex; the word ratios are 1.1, 2.6 to
3.0, 0.8 for Moshi, and 3.9.
The full-duplex models do not stop: 90\% of PersonaPlex's rollouts and
92\% of Moshi's were still speaking at the 30\,s cap. A harness audit found that reaching the cap does not by itself establish continued speech, so these durations are neither clean lower bounds nor single replies (\S\ref{sec:limitations}). Gemini reached the cap on 24\% of rollouts, GPT Realtime
on 2 to 3\%, Grok on 1\%. Similarity sits near its floor: median token F1
is 0.10 to 0.12 for every system that answers, against a floor of 0.08 and
a ceiling of 0.88. The systems respond to the conversation, and barely
more to the turn.

\begin{figure}[t]
\centering
\includegraphics[width=\linewidth]{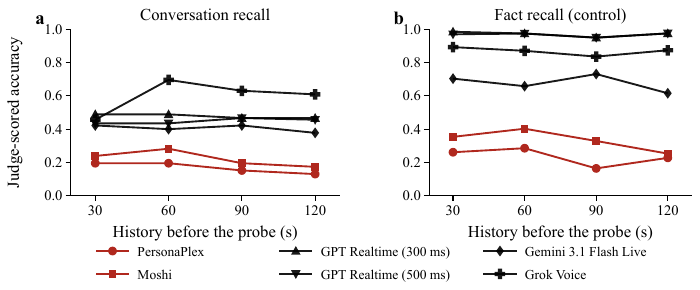}
\caption{Recall accuracy against history length, (a) conversation recall
and (b) fact recall, a general-knowledge control, judged by Claude Opus 5.5.
Full-duplex systems in red hear the history; hosted systems read it.}
\label{fig:recall}
\end{figure}

\textbf{Recall} (Figure~\ref{fig:recall}). Judged by Claude Opus 5.5 and given the history as text, Grok answers 46 to 70\% of questions about what was said, GPT Realtime 43 to 49\%, and Gemini 38 to 42\%, with no trend across history lengths from 30 to 120\,s. Hearing the same history as audio, PersonaPlex answers 13 to 20\% and Moshi 17 to 28\%. The fact-recall control separates memory from attention: GPT Realtime answers the general-knowledge question 95 to 98\% of the time, Grok 84 to 89\%, and Gemini 62 to 73\%, so their conversation-recall misses are misses of memory; PersonaPlex and Moshi answer it 16 to 40\% of the time, so most of theirs are failures to answer the question at all. Rescoring the same transcripts with GPT-4o moves no system by more than 9 points and preserves the ordering (Table~\ref{tab:judge}).

\section{Limitations}
\label{sec:limitations}

Hosted models read the history as text and full-duplex models hear it, so the cross-family recall gap is partly a gap between channels; streaming audio history to the hosted models is future work. The hosted systems' near-zero silence is a property of the silence-based endpointer configurations we tested, which request a response once speech ends, not a decision their models failed to make. Semantic turn detection was tested at one setting, low eagerness, on pause points only. A harness audit found that scoring could stall the shared audio stream and the drain could outlast queued playback, so reaching the 30\,s cap does not establish continued speech, full-duplex durations can include later resumptions, and latencies can reflect stalls. GPT-4o still selects the points and extracts the recall facts, sharing a provider with two systems under test. Turn-end comparisons depend on the family-specific response triggers of \S\ref{sec:protocol}; reply playback remains untested.
The gate is an energy threshold, no score has a human-human variance, and
we report one rollout per point. Labels are rule-based and only spot-checked: on 42 re-annotated windows, 86 to 95\% of boundaries fall within 250\,ms of a human's, a 250\,ms shift can reorder individual systems but not families (Appendix~\ref{app:extended}), and humans hear a backchannel in 7 of 20 pause windows labelled silent. Finally, the corpus is English and dyadic.

\section*{Ethics statement}

The audio is Seamless Interaction \citep{seamless}, released by Meta under
CC BY-NC 4.0. Its report states that paid adult participants gave informed consent for voice and image and were told to avoid personal details; flagged content was removed by review and the consent form is unpublished. We redistribute labels, point
identifiers, timings, and system outputs under the same licence, and no
audio or participant metadata. Recall probes use a participant's cloned
voice as benchmark input only; the clone is never released or presented as
the person. OpenAI and xAI assign
output ownership to the customer and Google disclaims it; none restricts
research publication. Released outputs are labelled machine-generated and
may not be used to train models. No new human-subject data was collected. \\


\FloatBarrier
\begin{ack}
This work was funded entirely by People Make Things, Inc.; no external grants or third-party funding supported it. The authors declare no other competing interests. We thank the Seamless Interaction team for releasing the corpus.
\end{ack}

\bibliographystyle{unsrtnat}

\appendix

\section{Selection and curation}
\label{app:selection}

\begin{figure}[H]
\centering
\includegraphics[width=\linewidth]{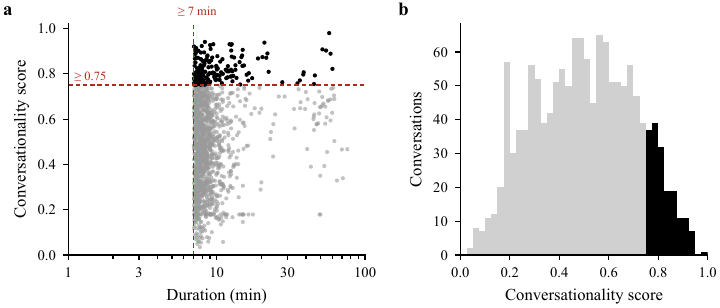}
\caption{Corpus selection. (a) Conversationality score against duration
for the 1{,}395 naturalistic conversations of at least 7 minutes; the 176
above 0.75 (black) form the corpus. (b) The score distribution, selected
conversations in black.}
\label{fig:selection}
\end{figure}

\textbf{Selection.} Seamless Interaction's naturalistic subset has
1{,}395 conversations of at least 7 minutes. The conversationality score
of \S\ref{sec:corpus} is the mean percentile rank of six per-conversation
statistics: speaking-time balance, speaker alternations per minute, share
of short utterances, question rate, one minus the longest monologue's
share, and overlaps per minute. Figure~\ref{fig:selection} shows the
score against duration; 176 conversations score at least 0.75 and are the
corpus. The cutoff is a choice about what to test, not a quality filter: a
conversation below it is a monologue with an audience, and a system's
turn-taking cannot be scored where the person never took a turn.

\textbf{Curation.} Table~\ref{tab:funnel} is the funnel from the labels to
the point set. The label-only gates can be recomputed from the released
labels: a pause point needs a labelled pause of at least 600\,ms, at least
1\,s of the speaker's turn before it, and no listener words inside it; a
turn-end point needs a user turn of at least 1\,s and three words and a
reply from the other speaker that starts within 3\,s and has at least five
words. The audio gates then require the pause window to gate as silence
(speech ratio at most 0.25) and the 3\,s before the boundary, and the
reply, to gate as speech (at least 0.2). Survivors are ranked by a
weighted sum of capped features (pause length, turn context, and an
ambiguity bit that is set unless the last word before the pause is a
conjunction, article, preposition, or filler), at most three per type per
conversation, and the GPT-4o judge re-ranks them; a judge score of zero is
never selected. The selected pauses have a median labelled duration of
0.96\,s (10th to 90th percentile 0.64 to 2.26\,s).

\begin{table}[H]
\caption{Curation funnel over the 176 conversations.}
\label{tab:funnel}
\centering
\small
\begin{tabular}{@{}l r r@{}}
\toprule
Stage & Pause points & Turn-end points \\
\midrule
Candidate boundaries in the labels & 26{,}094 & 19{,}933 \\
After label-only gates (length, words, adjacency, no listener words) & 8{,}068 & 7{,}029 \\
After audio gates, ranking cap (3 per type per conversation), and judge & 427 & 482 \\
Drawn for the run reported here (seed 0) & 300 & 300 \\
\bottomrule
\end{tabular}

\end{table}

\section{Label inventory and review}
\label{app:mapping}

\begin{figure}[H]
\centering
\includegraphics[width=\linewidth]{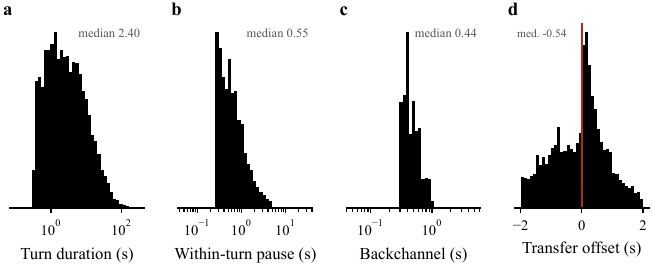}
\caption{Label distributions over the 176 conversations. (a) Turn
duration, (b) within-turn pause duration, (c) backchannel duration, all on
a log axis; (d) floor-transfer offset at speaker changes.}
\label{fig:labels}
\end{figure}

\textbf{Inventory.} Table~\ref{tab:inventory} counts the labels and
Figure~\ref{fig:labels} shows their distributions. Turns have a median
length of 2.4\,s and the median conversation has 9.4 turns, 2.2
backchannels, and 7.6 overlaps per minute. Within-turn pauses are short,
median 0.55\,s, which is why the 600\,ms gate removes most of them. The
floor-transfer offset in panel (d) is the next speaker's start minus the
previous speaker's end at each change of speaker; its median is negative
because an engine turn runs across the speaker's own pauses and the
listener's short responses begin before it ends.

\begin{table}[H]
\caption{Label inventory over the 176 conversations. Gap-fill words were
transcribed by Qwen3-ASR in stretches of at least 350\,ms that the source
transcripts left empty.}
\label{tab:inventory}
\centering
\small
\begin{tabular}{@{}l r r@{}}
\toprule
Item & Count & Per conversation (median) \\
\midrule
Conversations & 176 & -- \\
Hours & 35.9 & 8.7 min \\
Turns & 19{,}933 & 86 \\
Within-turn pauses & 26{,}094 & 114 \\
Backchannels & 4{,}974 & 22 \\
Overlaps & 16{,}698 & 72 \\
Words & 412{,}621 & 1750 \\
Words from gap-fill ASR & 18{,}552 (4.5\%) & -- \\
\midrule
Flags: gap-fill turn & 2{,}348 & -- \\
Flags: VAP says hold & 1{,}691 & -- \\
Flags: micro-turn & 769 & -- \\
Flags: VAP says shift & 639 & -- \\
\bottomrule
\end{tabular}

\end{table}

\textbf{Review.} The labels used for the evaluation points are the
engine's, and every evaluated window passed the two-annotator
accept-or-reject check of \S\ref{sec:labels}. Separately, four annotators
are re-annotating conversations in full in a review tool, and an
administrator approves each conversation.
At submission, 11 conversations (143 minutes, 1{,}260 engine turns) are
approved; Table~\ref{tab:review} compares them with the engine. We match
each engine turn to the human turn of the same speaker that overlaps it
most, and call it matched if the overlap covers more than half of it.
Three things follow. First, humans merge: 984 of the 1{,}260 engine turns
match a human turn, and the humans' 928 turns are fewer than the engine's.
Second, humans call more backchannels: 226 engine turns (18\%) became
backchannels, because the engine's rule (at most 1\,s and two words) is
stricter than a person's judgement. Third, boundaries move: of the matched
turns, 49\% have a start or end moved by more than 100\,ms, 39\% by more
than 250\,ms, and 22\% by more than 1\,s, the last being merges. The
flagger is informative but not sufficient: 81\% of
flagged engine turns were changed, against 55\% of
unflagged ones. For the benchmark these differences cut in a conservative
direction. A listener response the engine called a turn blocks a pause
point (a listener word inside the hold), so the mislabel removes points
rather than corrupting them, and a reply that a person would call a
backchannel is already excluded from turn-end points by the five-word
gate. The corrected labels will replace the engine's when the review is
complete, and every point will be regenerated from them.

\begin{table}[H]
\caption{Human review of 11 conversations against the engine labels.
Matched: engine turns whose overlap with a human turn of the same speaker
exceeds half their length. Moved: matched turns with a boundary shifted by
more than 100\,ms. Dropped: engine turns with no match, almost all
relabelled as backchannels. Flags: engine turns flagged for review.}
\label{tab:review}
\centering
\footnotesize
\setlength{\tabcolsep}{4pt}
\begin{tabular}{@{}l r r r r r r r r r@{}}
\toprule
Conversation & min & \multicolumn{2}{c}{turns} & backch. & pauses & matched & moved & dropped & flags \\
\cmidrule(lr){3-4}
 & & engine & human & human & human & & $>$100\,ms & & \\
\midrule
V00\_S1854\_I00000180 & 8.5 & 75 & 59 & 42 & 119 & 68 & 40 & 7 & 20 \\
V00\_S1966\_I00000215 & 8.4 & 69 & 47 & 41 & 151 & 60 & 39 & 9 & 26 \\
V03\_S0106\_I00000139 & 10.9 & 105 & 70 & 84 & 170 & 82 & 45 & 23 & 6 \\
V03\_S0165\_I00000506 & 8.9 & 109 & 107 & 43 & 102 & 86 & 55 & 23 & 21 \\
V03\_S0265\_I00000503 & 7.8 & 94 & 97 & 32 & 108 & 92 & 55 & 2 & 15 \\
V03\_S0320\_I00000060 & 7.9 & 90 & 58 & 61 & 216 & 67 & 41 & 23 & 28 \\
V03\_S0480\_I00000058 & 12.5 & 153 & 115 & 95 & 221 & 113 & 21 & 40 & 41 \\
V03\_S1094\_I00000197 & 12.5 & 63 & 43 & 129 & 169 & 48 & 29 & 15 & 20 \\
V03\_S1600\_I00000536 & 8.9 & 64 & 44 & 54 & 260 & 46 & 21 & 18 & 19 \\
V03\_S2060\_I00000196 & 20.3 & 238 & 125 & 173 & 363 & 154 & 80 & 84 & 90 \\
V03\_S2086\_I00000218 & 36.3 & 200 & 163 & 297 & 571 & 168 & 58 & 32 & 40 \\
\midrule
Total & 143.0 & 1260 & 928 & 1051 & 2450 & 984 & 484 & 276 & 326 \\
\bottomrule
\end{tabular}

\end{table}

\section{System cards}
\label{app:cards}
Table~\ref{tab:cards} lists the systems. All six ran on 10 September
2026, 08:45 to 12:23 PDT, from one Modal job
(\texttt{people-bench-eval}, T4 shards, one per provider) that read points
from S3 and mirrored every rollout back; Grok's turn-end and recall points
were rerun the same day, 17:22 to 18:13 PDT, under the corrected protocol
described below. The semantic turn-detection configuration ran on 4 October 2026 on the same harness, and all recall transcripts were rescored with Claude Opus 5.5 on 4 October 2026. Every
system received the same identity prompt frame (\S\ref{sec:protocol}), the
same 164\,s history cap, one scored rollout per point (a second is released), and 24\,kHz PCM16 audio.
After the streamed pause every system receives 3\,s of silence frames, plus
up to 8\,s more if it has produced no audio; at a turn-end point a
full-duplex model is streamed silence until it stops, a hosted model's
buffer is committed and a response requested with server VAD off, and either
way the system is stopped at 30\,s.
Timeouts and no-audio counts are in Table~\ref{tab:main} and \S\ref{sec:results}.
Grok's turn-end and recall rows come from a rerun under a corrected commit
protocol: xAI processes \texttt{input\_audio\_buffer.commit} asynchronously,
and a \texttt{response.create} sent in the same instant is answered before
the audio item exists, from the text history alone. The harness now waits
for the server's \texttt{committed} event plus 0.5\,s before requesting the
response, and disables xAI's VAD with \texttt{\{"type": null\}}, the only
form the endpoint accepts. Grok's turn-end latency therefore includes that
wait and overstates the model's own latency by roughly half a second.
The Gemini API was used on the paid tier, on which Google does not use
prompts or responses to improve its products.

\begin{table}[H]
\caption{System cards. Hosted VAD settings apply to pause points only; at
turn-end and recall points server VAD is off and the buffer is committed.}
\label{tab:cards}
\centering
\scriptsize
\setlength{\tabcolsep}{3pt}
\begin{tabular}{@{}>{\raggedright\arraybackslash}p{2.0cm} >{\raggedright\arraybackslash}p{3.2cm} >{\raggedright\arraybackslash}p{2.8cm} >{\raggedright\arraybackslash}p{1.6cm} >{\raggedright\arraybackslash}p{3.4cm}@{}}
\toprule
System & Model & Endpoint / host & History & Endpointer \\
\midrule
PersonaPlex & PersonaPlex-7B (NVIDIA) & self-hosted, Modal A100, 48 sessions & audio, two channels & none (full-duplex) \\
Moshi & \texttt{kyutai/moshiko-} \texttt{pytorch-bf16} & self-hosted, Modal A100, 48 sessions & audio, two channels & none (full-duplex) \\
GPT Realtime (300\,ms) & \texttt{gpt-realtime-} \texttt{2025-08-28}, voice alloy & OpenAI Realtime WebSocket, 16 sessions & text + trailing audio & server VAD, threshold 0.5, 300\,ms prefix, 300\,ms silence \\
GPT Realtime (500\,ms, default) & \texttt{gpt-realtime-} \texttt{2025-08-28}, voice alloy & OpenAI Realtime WebSocket, 16 sessions & text + trailing audio & server VAD, threshold 0.5, 300\,ms prefix, 500\,ms silence (OpenAI server-VAD default) \\
GPT Realtime (semantic, low) & \texttt{gpt-realtime-} \texttt{2025-08-28}, voice alloy & OpenAI Realtime WebSocket & text + trailing audio & semantic VAD, eagerness \texttt{low}; pause points only (475 points, 2 rollouts); API default eagerness is \texttt{auto} \\
Gemini 3.1 Flash Live & \texttt{gemini-3.1-flash-} \texttt{live-preview}, voice Kore & Gemini Live BidiGenerateContent, 16 sessions & text + trailing audio & automatic activity detection, 300\,ms prefix, 500\,ms silence \\
Grok Voice & \texttt{grok-voice-latest}, voice eve & xAI Realtime WebSocket, 8 sessions & text + trailing audio & server VAD, threshold 0.5, 300\,ms prefix, 300\,ms silence; off at turn ends via \texttt{\{"type": null\}} \\
\bottomrule
\end{tabular}
\end{table}

\section{Gate sensitivity}
\label{app:sens}

The speech gate has three parameters, threshold 0.01, release ratio 0.5,
and minimum duration 80\,ms, each recorded in every pause score so a score
can be recomputed from its own artifact. They were calibrated against
Whisper word timestamps over the systems' own output audio, a label source
independent of any energy threshold: each response is tiled with 2\,s
windows at a 1\,s hop, each window is peak-normalised before decoding so
the labeller cannot inherit the level bias being measured, and a word
counts only if Whisper's no-speech head, mean log-probability, and
repetition guard all pass and every window covering the instant agrees. On
human reference speech the labeller validates 49 of 62 windows, the misses
being real pauses; on digital silence, floor noise, broadband noise at 0.05,
and a 120\,Hz tone it labels nothing.

Table~\ref{tab:gate} gives the sweep. An Otsu split of the 21{,}580
envelope windows lands at $3.2\times10^{-4}$; against the labels it opens
intervals a median 840\,ms early and moves the pause score of the
calibration run from 0.544 to 0.228. Onset error keeps falling above 0.01
(mean absolute error 282\,ms at 0.01, 169\,ms at 0.05) but only by dropping
labelled speech; 0.01 is the highest threshold that still matches every
labelled utterance and the one at which the onset bias crosses zero
(median 0\,ms on pause rollouts, $-140$\,ms on turn-end rollouts). A missed
utterance is the error that matters, because it turns a takeover into a
hold. The same rule fixes the release ratio: 0.75 improves onset error
(308 against 410\,ms) but loses a labelled utterance. Minimum durations
from 40 to 80\,ms are indistinguishable. The gate remains a mediocre
classifier of lexical speech (AUC 0.80 on pause rollouts, 0.85 on
turn-end rollouts): labelled speech runs down to $8\times10^{-4}$ and
unlabelled audio up to $1.8\times10^{-2}$, which is the wordless phonation
of \S\ref{sec:metrics}, speech-like sound (dominant frequency 110 to
135\,Hz) that Whisper cannot decode and that is not an echo of the user
channel.

\begin{table}[H]
\caption{Gate sensitivity on the calibration run (57 PersonaPlex pause
rollouts with Whisper word labels).}
\label{tab:gate}
\centering
\small
\begin{tabular}{@{}l r r l@{}}
\toprule
Threshold & Pause score & Onset error & Labelled utterances missed \\
\midrule
$3.2\times10^{-4}$ (Otsu) & 0.228 & opens 840\,ms early (median) & 0 \\
0.01 (used) & 0.544 & 282\,ms MAE, bias 0 & 0 \\
0.05 & -- & 169\,ms MAE & $>$0 \\
\addlinespace[2pt]
Release ratio 0.75 (at 0.01) & -- & 308 vs 410\,ms MAE & 1 \\
Minimum duration 40--80\,ms & -- & no change & 0 \\
\bottomrule
\end{tabular}
\end{table}

\section{Extended results}
\label{app:extended}

\begin{figure}[H]
\centering
\includegraphics[width=0.62\linewidth]{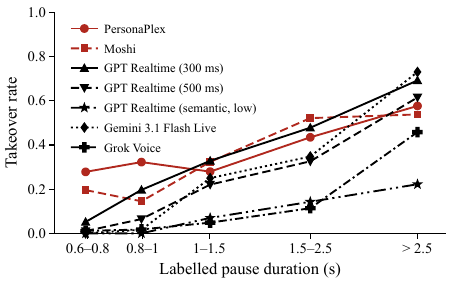}
\caption{Takeover rate against labelled pause duration; bins hold at least
ten rollouts. GPT Realtime (semantic, low) uses the 173 shared pause points.}
\label{fig:pausedur}
\end{figure}

\textbf{Takeover depends on the pause length.} Figure~\ref{fig:pausedur}
bins the pause points by labelled duration. Every system takes long pauses
more often than short ones. For the hosted systems the curve is their
latency read backwards: GPT Realtime takes 5\% of pauses under 0.8\,s and
69\% of pauses over 2.5\,s, because a response that arrives a second after
the pause began lands inside a long pause and after a short one. The
full-duplex models start higher, 20 to 28\% on the shortest pauses, and
rise less.

\begin{figure}[H]
\centering
\includegraphics[width=\linewidth]{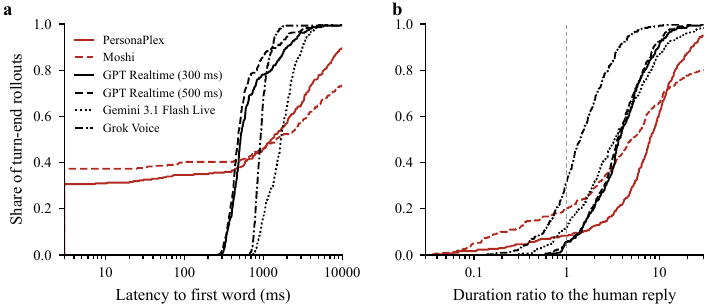}
\caption{Turn-end points, cumulative distributions over rollouts, log
axes. (a) Latency to first word, non-positive values drawn at the left
edge. (b)
Duration ratio to the human reply; the dashed line is the human.}
\label{fig:eotecdf}
\end{figure}

\textbf{Turn-end distributions.} Figure~\ref{fig:eotecdf} gives the full
distributions behind Figure~\ref{fig:eot}. Latency is bimodal by family:
the hosted systems are tight (GPT Realtime's interquartile range is under
100\,ms wide), and the full-duplex models have a mass at zero, words
already under way at the turn end, and a long tail. The duration ratio
shows what the medians hide: a third of Grok's replies are shorter than
the human's, almost none of PersonaPlex's, and Moshi's curve starts at zero
because a fifth of its rollouts contain no words.

\textbf{Recall at every history length.} Table~\ref{tab:recallall} gives
the accuracies behind Figure~\ref{fig:recall}. No system's conversation
recall falls monotonically with history length within 30 to 120\,s; the
differences between lengths are within the bootstrap intervals of
Table~\ref{tab:main}. Table~\ref{tab:exrecall} shows one point. The
question is asked in the user's cloned voice about something the user said
a minute earlier. Only Moshi answers correctly from the person's role; Gemini answers in the first person but gives the wrong fact, GPT Realtime answers as a counsellor to the user, which the identity prompt told it not to be, and Grok echoes the question's leaked tail without
answering and is still credited by the judge, an instance of the leniency
measured below.

\begin{table}[H]
\caption{Recall accuracy (\%) by history length, Claude Opus 5.5 judge.}
\label{tab:recallall}
\centering
\small
\begin{tabular}{@{}l rrrr rrrr@{}}
\toprule
 & \multicolumn{4}{c}{Conversation recall} & \multicolumn{4}{c}{Fact recall (control)} \\
\cmidrule(lr){2-5}\cmidrule(lr){6-9}
System & 30\,s & 60\,s & 90\,s & 120\,s & 30\,s & 60\,s & 90\,s & 120\,s \\
\midrule
PersonaPlex & 20 & 20 & 15 & 13 & 26 & 29 & 16 & 23 \\
Moshi & 24 & 28 & 20 & 17 & 35 & 40 & 33 & 25 \\
\addlinespace[2pt]
GPT Realtime (300 ms) & 49 & 49 & 47 & 47 & 98 & 97 & 95 & 97 \\
GPT Realtime (500 ms, default) & 43 & 43 & 47 & 46 & 97 & 97 & 95 & 97 \\
Gemini 3.1 Flash Live & 42 & 40 & 42 & 38 & 70 & 66 & 73 & 62 \\
Grok Voice & 46 & 70 & 63 & 61 & 89 & 87 & 84 & 87 \\
\bottomrule
\end{tabular}

\end{table}

\begin{table}[H]
\caption{One conversation-recall point, every system's reply.}
\label{tab:exrecall}
\centering
\footnotesize
\begin{tabular}{@{}>{\raggedright\arraybackslash}p{2.4cm} >{\raggedright\arraybackslash}p{9.9cm} c@{}}
\toprule
System & Spoken answer (Whisper transcript) & Score \\
\midrule
\multicolumn{3}{@{}p{12.9cm}@{}}{\textbf{Question} (60\,s of history): So, like, do you, um, regret messing with this girl, or what? For sure, for sure.} \\
\multicolumn{3}{@{}p{12.9cm}@{}}{\textbf{Gold answer}: I do regret messing with this girl.} \\
\midrule
PersonaPlex & I don't know. I mean, like, you know, like, you had the little thing going on. And you know, you kind of regretted it. And you know, it's like, I don't know. Like, maybe she, like, moved to another country with another guy, and you're like, ah, you know, that' & 0 \\
Moshi & It's not something I'm ashamed of. It's just something I regret. You know that there's there's there's a lot of things. I mean, I think the thing is is. & 1 \\
GPT Realtime (300 ms) & Man, for real, I'm not even gonna lie. It does sound like you're carrying some regret. But hey, it's all part of the game, right? We live and we learn. What's got you feeling that way about this situation, though? & 0 \\
GPT Realtime (500 ms, default) & Man, for real, it sounds like you're wrestling with some regret, huh? Like looking back, do you feel it was more trouble than it was worth? Or is it just one of those things where you're like, dang, I really should have known better. & 0 \\
Gemini 3.1 Flash Live & Nah, no regrets, man. You know how it is. Some people come into your life for a reason, some for a season. You know, learn a lesson. That kind of thing. What about you? & 0 \\
Grok Voice & for sure for sure. I mean, we all got that one, you know? What made you regret it? & 1 \\
\bottomrule
\end{tabular}

\end{table}

\textbf{Endpointer, paired.} GPT Realtime's two settings ran on the same
300 pause points. On the 296 points where both were scored, 32 took the
pause at both settings, 46 at 300\,ms only, 18 at 500\,ms only, and 200 at
neither. The longer silence window converts a takeover into a hold on one
point in six and the reverse on one in sixteen.

\textbf{Label verification on the windows in use.} All 600 pause and
turn-end windows were listened to and accepted by two annotators before the
run, without a per-window log. The measurable check is the 42 that fall in
the 11 conversations with an approved human re-annotation
(Appendix~\ref{app:mapping}). For the 20 pause
windows the nearest human pause of the same speaker starts within 100\,ms
of the engine's on 85\% and within 250\,ms on 90\% (median shift 0); its
end is within 250\,ms on 95\%. For the 22 turn-end windows the human turn
end is within 100\,ms on 82\% and within 250\,ms on 86\%, and the human
reply starts within 3\,s on 19. The one large disagreement is a pause the
humans place 6.7\,s later. The audio gates in curation are why these
windows agree far better than the corpus-wide 39\%: a selected pause must
gate as silence and its lead-in as speech. The humans also mark a listener
backchannel inside 7 of the 20 pause windows where the engine found no
listener words; the person held the floor at those points, but the other
person was not silent.

\textbf{Boundary sensitivity.} Table~\ref{tab:perturb} rescores the
takeover rate with the labelled pause end moved by $\pm$250\,ms and the
start by $+$250\,ms, from the stored onsets. The hosted rates move the most
when the end moves, because their first words land near it: GPT Realtime
at 300\,ms goes from 0.15 to 0.40 across the end shifts and overtakes
PersonaPlex at $+$250\,ms. The full-duplex rates move the most when the
start moves, because a share of their takeovers begins at or before it. A
250\,ms shift can therefore reorder individual systems by takeover rate;
what no shift changes is that the hosted systems never stay silent, that
their words land after the pause on most points, that Grok takes the fewest
pauses, and the sign of the GPT Realtime endpointer effect.

\begin{table}[H]
\caption{Takeover rate under boundary perturbation.}
\label{tab:perturb}
\centering
\small
\begin{tabular}{@{}l r r r r@{}}
\toprule
System & as labelled & pause end $-250$\,ms & pause end $+250$\,ms & pause start $+250$\,ms \\
\midrule
PersonaPlex & 0.34 & 0.29 & 0.36 & 0.15 \\
Moshi & 0.29 & 0.24 & 0.31 & 0.18 \\
\addlinespace[2pt]
GPT Realtime (300 ms) & 0.26 & 0.15 & 0.40 & 0.26 \\
GPT Realtime (500 ms, default) & 0.17 & 0.12 & 0.28 & 0.17 \\
Gemini 3.1 Flash Live & 0.18 & 0.07 & 0.21 & 0.17 \\
Grok Voice & 0.07 & 0.03 & 0.09 & 0.07 \\
\bottomrule
\end{tabular}

\end{table}

\textbf{Judge check.} Recall was first judged by GPT-4o while two GPT systems were under test; the reported scores come from rescoring every transcript with Claude Opus 5.5, which agrees with GPT-4o on 97 to 99\% of control rollouts and 82 to 93\% of conversation-recall rollouts, moving no system by more than 9 points (PersonaPlex 8 to 17, Grok 57 to 60) and preserving the ordering. Table~\ref{tab:judge} also scores the same transcripts with a strict
matcher that needs no model: for fact recall, every content word of the
gold answer must appear in the transcript; for conversation recall, at
least half of them. On the control the two scorers agree on 95 to 99\% of
rollouts for every system. On conversation recall GPT-4o and the matcher agree on 79 to 84\%
for the full-duplex models but on 48 to 63\% for the hosted ones, whose
replies paraphrase: GPT-4o credits Grok on 76 replies and GPT Realtime
on 59 to 63 that the matcher rejects (the J-only column), against 12 for
PersonaPlex. Under the strict matcher the hosted systems tie at 19 to 26\%
(Grok 26\%, GPT Realtime 23\%, Gemini 19\%) ahead of PersonaPlex at 15\%
and Moshi at 12\%; the hosted lead over the full-duplex models shrinks by about two thirds under either model judge and
Grok's lead over GPT Realtime shrinks to three points. Both judges' leniency is toward paraphrase, which the hosted systems produce; a reader who distrusts model judges can use the strict column. For point selection the judge's
verdicts are not in the released manifest, so Table~\ref{tab:tercile} bins
pause points by the heuristic score that ranked them: GPT Realtime's
takeover rate is below the full-duplex models' in the lowest third and
level with them in the highest, so the selection did not favour it.

\begin{table}[H]
\caption{Recall accuracy (\%) under the Claude Opus 5.5 judge, the original GPT-4o judge, and a strict word
matcher over the same scored rollouts, GPT-4o/matcher agreement, and the number of
rollouts credited by only one of those two (J only: GPT-4o; S only: matcher).}
\label{tab:judge}
\centering
\footnotesize
\setlength{\tabcolsep}{3.5pt}
\begin{tabular}{@{}l rrrrrr rrrrrr@{}}
\toprule
 & \multicolumn{6}{c}{Fact recall (control)} & \multicolumn{6}{c}{Conversation recall} \\
\cmidrule(lr){2-7}\cmidrule(lr){8-13}
System & Opus & GPT-4o & strict & agree & J only & S only & Opus & GPT-4o & strict & agree & J only & S only \\
\midrule
PersonaPlex & 23 & 23 & 27 & 95\% & 1 & 13 & 17 & 8 & 15 & 79\% & 12 & 26 \\
Moshi & 33 & 31 & 32 & 97\% & 3 & 6 & 22 & 20 & 12 & 84\% & 22 & 8 \\
\addlinespace[2pt]
GPT Realtime (300 ms) & 97 & 96 & 96 & 99\% & 2 & 2 & 48 & 46 & 23 & 58\% & 59 & 17 \\
GPT Realtime (500 ms, default) & 97 & 96 & 96 & 99\% & 2 & 0 & 45 & 46 & 23 & 54\% & 63 & 22 \\
Gemini 3.1 Flash Live & 68 & 67 & 66 & 96\% & 9 & 4 & 41 & 37 & 19 & 63\% & 49 & 17 \\
Grok Voice & 87 & 86 & 86 & 99\% & 1 & 2 & 60 & 57 & 26 & 48\% & 76 & 19 \\
\bottomrule
\end{tabular}

\end{table}

\begin{table}[H]
\caption{Takeover rate by tercile of the curation ranking score.}
\label{tab:tercile}
\centering
\small
\begin{tabular}{@{}l r r r@{}}
\toprule
System & lowest third & middle third & highest third \\
\midrule
PersonaPlex & 0.28 & 0.27 & 0.47 \\
Moshi & 0.23 & 0.19 & 0.48 \\
\addlinespace[2pt]
GPT Realtime (300 ms) & 0.07 & 0.23 & 0.49 \\
GPT Realtime (500 ms, default) & 0.03 & 0.10 & 0.38 \\
Gemini 3.1 Flash Live & 0.00 & 0.11 & 0.42 \\
Grok Voice & 0.01 & 0.03 & 0.18 \\
\bottomrule
\end{tabular}

\end{table}

\textbf{Ceilings.} A GPT-4o text model given the same transcript prefix and
question answers 97\% of the conversation-recall questions (100, 93, 98,
and 96\% at 30, 60, 90, and 120\,s), judged by GPT-4o with the same prompt (not rescored with Opus); the
questions are answerable, and the best live system reaches two thirds of
that.
For similarity, the human reply's own label transcript against its Whisper
transcript gives median token F1 0.88 and cosine 0.89 over the 300 turn-end
points; every system's median lies just above the shuffled-reply floor and
far below that ceiling.

\textbf{Confidence intervals.} Table~\ref{tab:ci} gives 95\% intervals
for every value in Table~\ref{tab:main}, from 2{,}000 bootstrap resamples
of conversations. The GPT Realtime endpointer difference in takeover rate is consistent across resamples, though the intervals overlap slightly (0.21 to 0.31 against 0.13 to 0.22); the two full-duplex models' takeover rates overlap substantially.

\begin{table}[H]
\caption{95\% bootstrap intervals over conversations for
Table~\ref{tab:main}.}
\label{tab:ci}
\centering
\footnotesize
\setlength{\tabcolsep}{4pt}
\begin{tabular}{@{}l r r r r r r@{}}
\toprule
 & \multicolumn{6}{c}{Pause points and recall at 120\,s} \\
\cmidrule(lr){2-7}
System & silent & takeover & after & takeover lat.\ (ms) & recall conv. & recall fact \\
\midrule
PersonaPlex & [0.17, 0.26] & [0.29, 0.39] & [0.34, 0.44] & [116, 357] & [0.04, 0.24] & [0.14, 0.33] \\
Moshi & [0.33, 0.45] & [0.24, 0.35] & [0.24, 0.35] & [242, 428] & [0.07, 0.28] & [0.16, 0.35] \\
\addlinespace[2pt]
GPT Realtime (300 ms) & [0.00, 0.02] & [0.21, 0.31] & [0.67, 0.77] & [905, 1117] & [0.33, 0.62] & [0.94, 1.00] \\
GPT Realtime (500 ms, default) & [0.00, 0.04] & [0.13, 0.22] & [0.76, 0.85] & [981, 1227] & [0.28, 0.57] & [0.94, 1.00] \\
GPT Realtime (semantic, low)$^{\dagger}$ & [0.02, 0.09] & [0.03, 0.10] & [0.83, 0.93] & [1108, 1528] & -- & -- \\
Gemini 3.1 Flash Live & [0.00, 0.00] & [0.14, 0.22] & [0.77, 0.86] & [1899, 2883] & [0.24, 0.51] & [0.50, 0.72] \\
Grok Voice & [0.00, 0.00] & [0.04, 0.10] & [0.81, 0.90] & [2158, 4258] & [0.48, 0.74] & [0.80, 0.95] \\
\bottomrule
\end{tabular}
\\[6pt]
\begin{tabular}{@{}l r r r r r r@{}}
\toprule
 & \multicolumn{6}{c}{Turn-end points} \\
\cmidrule(lr){2-7}
System & lat.\ (ms) & cens. & dur.\ ratio & word ratio & token F1 & cosine \\
\midrule
PersonaPlex & [745, 1560] & [0.86, 0.93] & [7.00, 8.81] & [3.08, 4.67] & [0.09, 0.11] & [0.15, 0.18] \\
Moshi & [0, 819] & [0.89, 0.95] & [2.76, 4.20] & [0.61, 1.00] & [0.00, 0.06] & [0.10, 0.14] \\
\addlinespace[2pt]
GPT Realtime (300 ms) & [495, 551] & [0.01, 0.05] & [3.40, 4.09] & [2.84, 3.29] & [0.33, 0.60] & [0.94, 1.00] \\
GPT Realtime (500 ms, default) & [454, 501] & [0.01, 0.04] & [3.24, 3.96] & [2.64, 3.36] & [0.33, 0.61] & [0.94, 1.00] \\
Gemini 3.1 Flash Live & [1580, 1771] & [0.19, 0.29] & [2.79, 3.95] & [2.26, 3.00] & [0.10, 0.13] & [0.16, 0.20] \\
Grok Voice & [897, 952] & [0.00, 0.02] & [1.22, 1.55] & [1.00, 1.29] & [0.11, 0.14] & [0.16, 0.19] \\
\bottomrule
\end{tabular}

\end{table}

\textbf{Failures.} Of 6{,}504 rollouts, 15 failed in transport and are
excluded; no-audio rollouts are 7\% of Grok's pause points and at most 1\%
elsewhere. Table~\ref{tab:agreement} lists every non-scored outcome per
system. Transport errors are 15 in total (WebSocket closes and
one timeout), listed per evaluation; they are retried on a restart of the
same run directory and excluded here.

\begin{table}[H]
\caption{Exceptional outcomes per system. Only errors are excluded from scoring; they are transport failures.
Censored: turn-end rollouts still speaking at the 30\,s cap. No words:
scored turn-end rollouts with no transcribed word.}
\label{tab:agreement}
\centering
\footnotesize
\setlength{\tabcolsep}{4pt}
\begin{tabular}{@{}l r r r r r r r@{}}
\toprule
 & \multicolumn{2}{c}{Pause} & \multicolumn{3}{c}{Turn end} & \multicolumn{2}{c}{Recall} \\
\cmidrule(lr){2-3}\cmidrule(lr){4-6}\cmidrule(lr){7-8}
System & no audio & errors & censored & no words & errors & no audio & errors \\
\midrule
PersonaPlex & 0 & 2 & 270 & 10 & 0 & 0 & 0 \\
Moshi & 0 & 1 & 277 & 57 & 0 & 0 & 0 \\
\addlinespace[2pt]
GPT Realtime (300 ms) & 3 & 1 & 9 & 0 & 0 & 5 & 0 \\
GPT Realtime (500 ms, default) & 0 & 0 & 6 & 0 & 1 & 3 & 1 \\
GPT Realtime (semantic, low)$^{\dagger}$ & 0 & 0 & -- & -- & -- & -- & -- \\
Gemini 3.1 Flash Live & 2 & 0 & 71 & 0 & 0 & 0 & 9 \\
Grok Voice & 20 & 0 & 3 & 0 & 0 & 0 & 0 \\
\bottomrule
\end{tabular}

\end{table}

\clearpage

\end{document}